%% file: main.tex
\documentclass[sigconf,10pt]{acmart}

\AtBeginDocument{%
  \setlength{\textfloatsep}{6pt plus 1pt minus 2pt}
  \setlength{\floatsep}{4pt plus 1pt minus 2pt}
  \setlength{\intextsep}{6pt plus 1pt minus 2pt}

}
\definecolor{sectiongray}{gray}{0.85} 
\definecolor{tiergray}{gray}{0.92}    

\AtBeginDocument{%
  }

\usepackage{graphicx}
\usepackage{booktabs}
\usepackage{makecell}
\usepackage{pifont}
\usepackage{multirow}
\usepackage{textcase}
\usepackage{enumitem}
\usepackage[table]{xcolor}
\usepackage{subcaption}
\newcommand{\rev}[1]{#1}

\makeatletter
\def\acm@section@titlefont{\scshape\MakeTextUppercase}
\def\acm@subsection@titlefont{\scshape\MakeTextUppercase}
\def\acm@subsubsection@titlefont{\scshape\MakeTextUppercase}

\makeatother
\makeatletter

\renewcommand{\abstractname}{\MakeTextUppercase{Abstract}}
\renewcommand{\keywordsname}{\MakeTextUppercase{Keywords}}
\renewcommand{\refname}{\MakeTextUppercase{References}}
\renewcommand{\acksname}{\MakeTextUppercase{Acknowledgements}}

\patchcmd{\@sect}
  {#6\@@par}
  {#6\MakeTextUppercase\@@par}
  {}{}
\setcopyright{acmlicensed}
\copyrightyear{2026}
\acmYear{2026}
\setcopyright{cc}
\setcctype{by}
\acmConference[MobiCom '26]{The 32nd Annual International Conference on Mobile Computing and Networking}{October 26--30, 2026}{Austin, TX, USA}
\acmBooktitle{The 32nd Annual International Conference on Mobile Computing and Networking (MobiCom '26), October 26--30, 2026, Austin, TX, USA}
\acmPrice{}
\acmDOI{10.1145/3795866.3796695}
\acmISBN{979-8-4007-2505-0/2026/10}

\begin{document}

\title{IoT-Brain: Grounding LLMs for Semantic-Spatial Sensor Scheduling}







\author[Zhou et al.]{
  Zhaomeng Zhou\textsuperscript{1}, 
  Lan Zhang\textsuperscript{1,2,*}, 
  Junyang Wang\textsuperscript{1},   
  Mu Yuan\textsuperscript{3}, \\
  Junda Lin\textsuperscript{1}, 
  Jinke Song\textsuperscript{4}
}

\thanks{* Lan Zhang is the corresponding author.}

\affiliation{%
  \institution{\textsuperscript{1}University of Science and Technology of China}
  \institution{\textsuperscript{2}Institute of Artificial Intelligence, Hefei Comprehensive National Science Center}
  \institution{
    \textsuperscript{3}The Chinese University of Hong Kong \quad 
    \textsuperscript{4}The Hong Kong University of Science and Technology
  }
  \country{}
}

\affiliation{%
  \institution{
    \{zhouzhm, iswangjy, linjunda\}@mail.ustc.edu.cn, zhanglan@ustc.edu.cn \\
    muyuan@cuhk.edu.hk, jikesog@gmail.com
  }
  \country{}
}

\begin{abstract}
Intelligent systems powered by large-scale sensor networks are shifting from predefined monitoring to intent-driven operation, revealing a critical \textit{Semantic-to-Physical Mapping Gap}. While large language models (LLMs) excel at semantic understanding, existing perception-centric pipelines operate retrospectively, overlooking the fundamental decision of what to sense and when. We formalize this proactive decision as \textit{Semantic–Spatial Sensor Scheduling (S³)} and demonstrate that direct LLM planning is unreliable due to inherent gaps in representation, reasoning, and optimization. To bridge these gaps, we introduce the \textit{Spatial Trajectory Graph (STG)}, a neurosymbolic paradigm governed by a "verify-before-commit" discipline that transforms open-ended planning into a verifiable graph optimization problem.
Based on \textit{STG}, we implement \textit{IoT-Brain}, a concrete system embodiment, and construct \textit{TopoSense-Bench}, a campus-scale benchmark with 5,250 natural-language queries across 2,510 cameras. Evaluations show \textit{IoT-Brain} boosts task success rate by 37.6\% over the strongest search-intensive methods while running nearly 2$\times$ faster and using 6.6$\times$ fewer prompt tokens. In real-world deployment, it approaches the reliability upper bound set by reducing 4.1$\times$ network bandwidth, providing a foundational framework for LLMs to interact with the physical world with unprecedented reliability and efficiency.
\end{abstract}



\maketitle


\section{INTRODUCTION}
\label{sec:introduction}

\begin{figure*}[t]
    \centering
    \setlength{\abovecaptionskip}{0.03cm}
    \includegraphics[width=\textwidth]{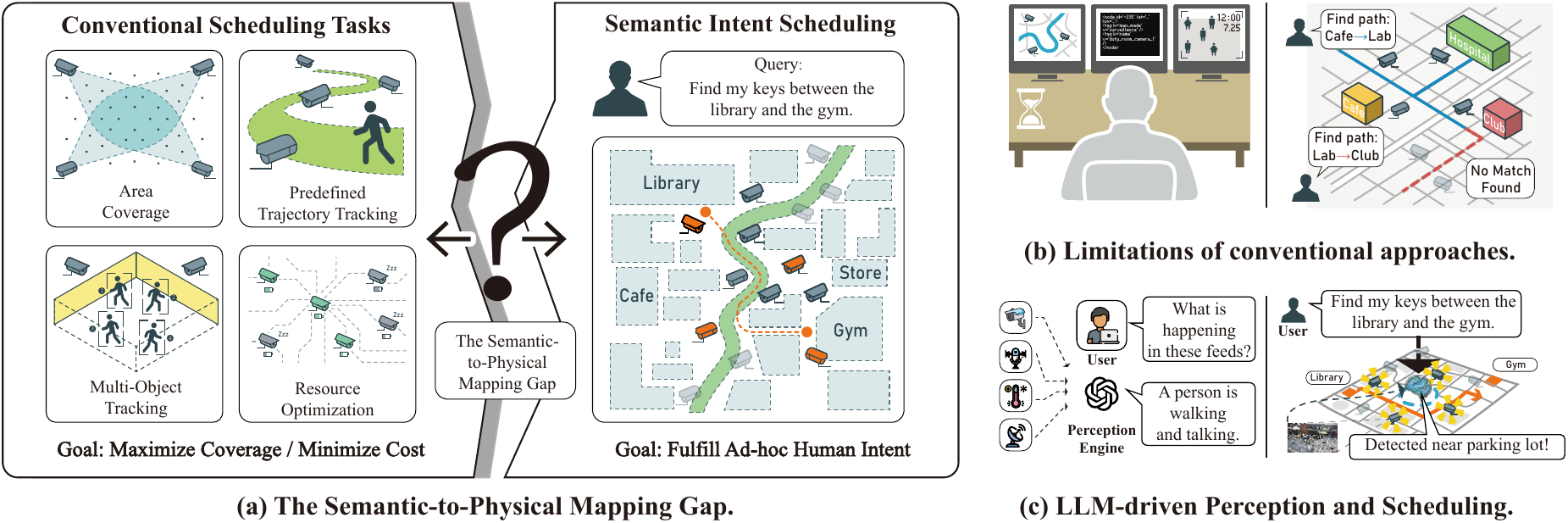}
    \caption{The challenge of the \textit{S\textsuperscript{3}} problem.}
    \label{fig:intro_problem}
    \vspace{-5mm}
\end{figure*}

The proliferation of large-scale sensor networks in smart cities and industries is catalyzing a paradigm shift towards intelligent automation\cite{Lin2017ASO,Menouar2017UAVEnabledIT,Peng2023DistributedIF}. This leap from traditional coverage-driven optimization\cite{Wu2024CooperativeSF, Cai2022MeMOTMT, Jia2023EnergyCM, Reily2020AdaptationTT} to real-time semantic goal satisfaction introduces a profound, yet underexplored challenge. A simple request like \textit{"Can you help me check my wallet between the library and the gym?"} exposes a stark divide between the vagueness of human language and the precise physical operations a sensor network must execute. We term this the \textit{Semantic-to-Physical Mapping Gap} (Fig.~\ref{fig:intro_problem}(a)), a fundamental hurdle that renders conventional task-specific models and rigid scripts ineffective.

The advent of large language models (LLMs) has provided a powerful engine for semantic interpretation~\cite{OpenAI2023GPT4TR, Meta2023Llama2,Google2025Gemini25,DeepSeekAI2025R1},  offering a promising path forward. Recent work on "Penetrative AI" further shows that LLMs can reason directly over raw sensor data~\cite{Xu2023PenetrativeAM, Yu2025SensorChatAQ, Ren2025TowardSL,Cheng2024AutoIoTAI}. Yet these advances largely remain within operates retrospectively, assuming the relevant sensor streams are already available, what we term \textit{Reactive Perception} (Fig.\ref{fig:intro_problem}(c)). This perspective overlooks a more fundamental upstream challenge, \textit{Proactive Scheduling}, that is paramount in large-scale deployments. Before any meaningful perception can occur, LLMs must decide what to sense and when to attend. The decision precedes and enables all subsequent perception but has received limited systematic study~\cite{king2024sasha, Liu2025TaskSenseAT, AlSafi2025VegaLI}. We formalize this pivotal challenge as the \textit{Semantic–Spatial Sensor Scheduling (S³)} problem.

Solving the \textit{S³} problem with off-the-shelf LLMs is far from straightforward. Our preliminary study~(\S\ref{subsec:preliminary_study}) tasks an LLM with end-to-end scheduling in a real-world topological environment, revealing three fundamental challenges:

\noindent\textbf{(1) Symbol-to-Semantic Chasm.} LLMs' native shortcoming in comprehending raw, machine-oriented symbolic topologies prevents them from building an effective world model, slashing their planning success by over 5$\times$ compared to when provided with structured, human-readable knowledge. 

\noindent\textbf{(2) Inferential Leap from Points to Paths.} LLMs' profound difficulty in inferring topological relationships like connectivity from disconnected symbols causes even a perfectly informed model to achieve merely 26\% trajectory coverage, leading to fragmented and unsound paths.

\noindent\textbf{(3) Optimization Shortfall in LLM Planning.} The inherent "satisficing" nature of LLMs leads to resource-heavy plans, exhibiting up to 45\% redundant sensor overlap even with structured guidance and compounding to a staggering 48$\times$ token overhead when processing raw symbolic data.

\textbf{Core idea: verify-before-commit.} This work aims to bridge the gap between an LLM's high-level semantic reasoning and the need for resource-efficient, physically grounded sensor scheduling. Our insight stems from observing LLMs’ native behavior on the \textit{S³} problem, showing that their plans are often ungrounded and glaringly misaligned with sensor-network constraints. By contrast, when supplied with pre-vetted, task-relevant topological knowledge, LLM-derived plans become both reliable and efficient. Such oracle-like access, however, is infeasible in dynamic deployments. In resource-sensitive environments, executing a speculative, unverified plan is prohibitively expensive. We therefore adopt a paradigm in which the LLM must proactively and autonomously discover and validate the necessary grounding knowledge through direct interaction with the physical world before any operational decision is taken. We term this disciplined approach "verify-before-commit", a principle requiring that all semantic hypotheses be fully validated against reality before they are deemed executable.

\textbf{STG design.} The "verify-before-commit" principle's progressive translation of high-level intent into concrete action inherently embodies a search for an optimal topological pathway within semantic constraints. This complex search process mirrors the established paradigm of graph construction\cite{Yang2021GraphbasedTE,Chen2024PlanonGraphSA,Wang2021StructuredSM}, which we formalize as the \textit{Spatial Trajectory Graph (STG)}, a neurosymbolic paradigm that operates through a systematic, multi-stage workflow.
The paradigm first mandates the structuring of ambiguous intent into a verifiable, hypothesized graph. It then requires grounding of the graph through an iterative validation loop against a physical world model, methodically transforming uncertainty into verified facts. Finally, the paradigm concludes with the optimization of the now-verified graph into a resource-aware, dynamic execution plan. This principled decomposition systematically separates semantic interpretation from deterministic validation and scheduling.

\begin{figure*}[t]
    \centering
    \setlength{\abovecaptionskip}{0.03cm}

    \begin{subfigure}[t]{0.49\textwidth}
        \centering
        \setlength{\abovecaptionskip}{0.cm}
        \includegraphics[width=\textwidth]{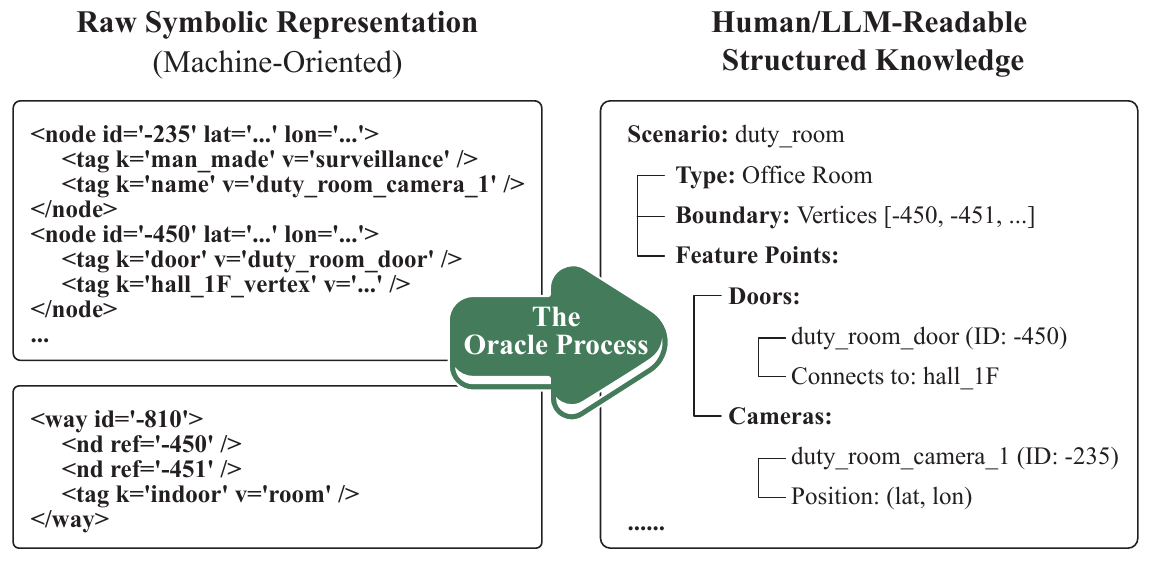}
        \caption{From raw symbols to structured knowledge.}
        \label{fig:representation_gap_sub}
    \end{subfigure}
    \hfill
    \begin{subfigure}[t]{0.49\textwidth}
        \centering
        \setlength{\abovecaptionskip}{0.cm}
        \includegraphics[width=\textwidth]{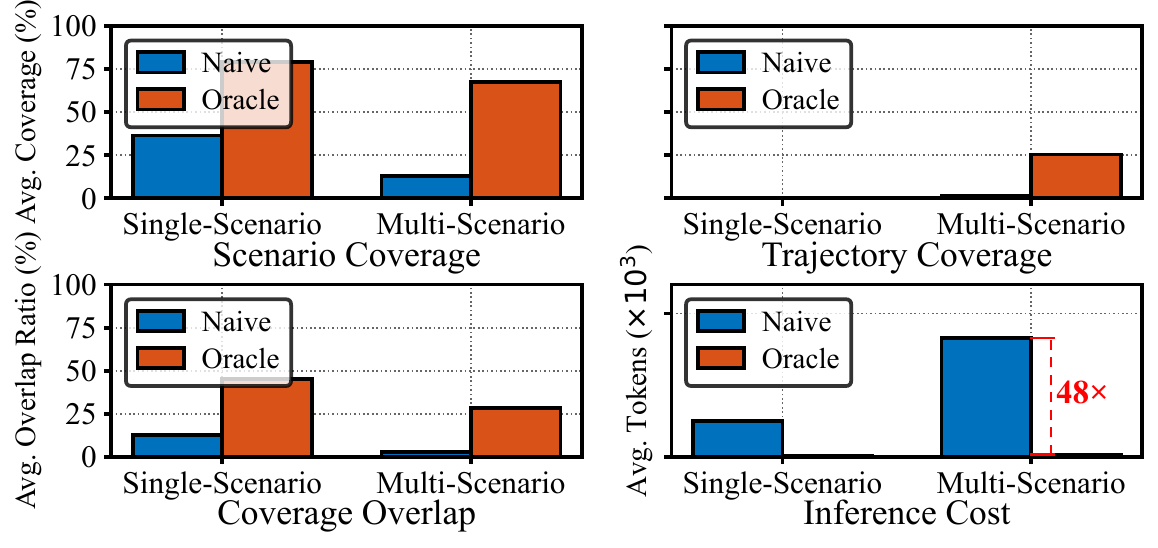} 
        \caption{Empirical planning failures.}
        \label{fig:prelim_results_sub}
    \end{subfigure}

    \caption{The representation gap and resulting planning failures.}
    \label{fig:representation_and_failures}
    \vspace{-6mm}
\end{figure*}

\textbf{IoT-Brain Implementation.} We implement the \textit{STG} paradigm in \textit{IoT-Brain}, a system engineered for robust, real-world interaction. Instead of a monolithic pipeline, \textit{IoT-Brain} adopts a modular architecture that leverages an LLM's advanced tool-calling and programming capabilities\cite{Qu2024ToolLW,Jiang2024ASO}. It employs a reactive, tool-based loop to continuously ground its semantic reasoning against our physical world model, effectively transforming abstract hypotheses into verifiable facts. This design strategically decouples the LLM's high-level semantic inference from the deterministic, resource-conscious tasks of scheduling and sensor control. The entire workflow is further accelerated by shared memory modules that cache executable reasoning evidence, amortizing the cost of interaction across sessions. To evaluate \textit{IoT-Brain}, we also constructed \textit{TopoSense-Bench}, a large-scale benchmark tailored to the \textit{S³} problem featuring a university-scale digital twin with 2,510 cameras and 5,250 real-world queries.

\textbf{Contributions} of our work are summarized as follows:

\begin{itemize}[leftmargin=*, topsep=0pt, partopsep=0pt, itemsep=0pt, parsep=0pt]
    \item We identify and formalize the \textit{S³} problem, a critical yet overlooked challenge in intent-driven sensor networks, and introduce the \textit{Spatial Trajectory Graph (STG)}, a neurosymbolic paradigm that grounds LLM reasoning in a verifiable, physical-world structure.

    \item We design and implement \textit{IoT-Brain}, our concrete system instantiation of \textit{STG}, and we construct and release \textit{TopoSense-Bench}, a large-scale benchmark with 5,250 real-world queries to catalyze future research in this domain.

    \item We conduct evaluations of \textit{IoT-Brain} on our \textit{TopoSense-Bench} and a physical testbed. Experimental results demonstrate the superiority of \textit{STG}. On the benchmark's most complex tasks, \textit{IoT-Brain} boosts task success by 37.6\% over the strongest search-intensive methods\cite{Qin2023ToolLLMFL,Chen2024SmurfsMS} while running nearly 2$\times$ faster and using 6.6$\times$ fewer prompt tokens. Furthermore, in live deployments, our system approaches the reliability of a resource-agnostic upper-bound approach while consuming 4.1$\times$ less network bandwidth.
\end{itemize}

\vspace{-0.06in}
\section{BACKGROUND \& MOTIVATION}
\label{sec:background}

We first survey the sensor-scheduling landscape, contrasting coverage-driven methods with emerging LLMs' capabilities to expose a critical gap~(\S\ref{subsec:need_for_semantic}). Then we present a preliminary study that decomposes LLM-based proactive scheduling into three fundamental gaps~(\S\ref{subsec:preliminary_study}). Next, we motivate a new planning paradigm that addresses these gaps~(\S\ref{subsec:motivation_paradigm}).

\vspace{-0.06in}
\subsection{Background}
\label{subsec:need_for_semantic}

Classical research in sensor networks has centered predominantly on coverage-driven optimization tasks, such as maximizing sensor coverage or tracking objects within a static, pre-defined field-of-view~\cite{Kumar2005BarrierCW, Chen2007DesigningLA, Cai2022MeMOTMT}. While effective for well-defined engineering objectives, this paradigm is fundamentally ill-suited for intent-driven queries specified in ambiguous natural language. In practice, such ad-hoc semantic tasks are often relegated to rudimentary solutions, typically relying on laborious human-in-the-loop operations~\cite{Hodgetts2017SeeNE, Pelletier2015AtypicalVD,Marois2020RealTimeGC} or brittle fixed-topology scripts~\cite{Brackenbury2019HowUI, Huang2015SupportingMM}. As illustrated in Fig.~\ref{fig:intro_problem}(b), both approaches break down in large-scale, dynamic settings~\cite{Tang2019CityFlowAC,Shim2021MultiTargetMV,Wu2021AMV}. Manual operation does not scale~\cite{Donald2015WorkEA}, and hard-coded scripts fail to generalize to the fluid and unpredictable nature of human intent~\cite{Ur2016TriggerActionPI, Morgan2022ReducingRT}.

The emergent semantic understanding of LLMs offers a transformative and promising route beyond the historical constraints of traditional sensor networks. Pioneering systems show LLMs can parse complex sensor data and interact with the physical world, opening a new frontier for AIoT~\cite{Cheng2024AutoIoTAI, Gao2024ChatIoTZG, Shen2025GPIoTTS,Ren2025TowardSL,Yu2025SensorChatAQ}. Yet, the current LLM-for-AIoT landscape is structurally imbalanced. As shown in Fig.~\ref{fig:intro_problem}(c), most work concentrates on \textit{Reactive Perception}, where models passively analyze pre-collected sensor streams. We instead tackle \textit{Proactive Scheduling}, the upstream problem of deciding precisely which sensors to activate and when. This shift from passive interpretation to active, goal-directed participation is essential for truly autonomous AIoT systems, and the transition is nontrivial, exposing fundamental challenges in reliably grounding LLM reasoning in the physical world. 

\vspace{-0.16in}
\subsection{Preliminary Study}
\label{subsec:preliminary_study}

To dissect why the leap from perception to scheduling is challenging for LLMs, we conducted a preliminary study designed to answer a core question. \textit{Can LLMs effectively plan routes and schedule sensors when given raw, symbolic topological data?} To investigate this, we constructed a dedicated testbed using real-world topological data from OpenStreetMap (OSM)~\cite{Haklay2008OpenStreetMapUS}, encompassing five multi-scenario buildings and a set of 373 manually curated queries that require both spatial understanding and path planning.

We compared two approaches. In the Naive setting, the LLM is prompted directly with machine-oriented OSM text (e.g., XML-style nodes and ways), and is asked to produce a sensor activation plan. In the Oracle setting, which serves as an upper bound, the same symbolic data is preprocessed into a structured, human-readable knowledge base that makes locations, connections, and salient features explicit (see Fig.~\ref{fig:representation_and_failures}(a)) for which the LLM then uses for reasoning. We assessed each approach on its ability to produce correct and efficient schedules, measuring scenario coverage, trajectory coverage, resource overlap, and token consumption. The results, summarized in Fig.~\ref{fig:representation_and_failures}(b), reveal three tightly coupled gaps that together impede reliable LLM-based scheduling.

\noindent\textbf{Gap 1: Symbol-to-Semantic Chasm.} The disparity between the Naive and Oracle settings reveals a fundamental representation mismatch.
The quantitative impact of this chasm is stark. When the LLM is provided with structured, human-readable knowledge, scenario coverage boosts by over 2.1$\times$ on single-scenario tasks and a remarkable 5$\times$ on multi-scenario tasks compared to the baseline Naive approach.
Because LLMs are trained on natural language rather than machine-oriented symbolic topologies, they struggle to effectively parse OSM-style structures~\cite{Feng2024CityGPTEU,Feng2025UrbanLLaVAAM,Sabbata2025GeospatialMI} and therefore cannot assemble a usable world model.
Planning performance consequently collapses, underscoring the need for a dedicated symbol-to-semantic translation layer.

\noindent\textbf{Gap 2: Inferential Leap from Points to Paths.}
Trajectory coverage lays bare a deeper reasoning failure.
In the Naive setting, it is essentially zero.
Even with an explicit, intelligible map, the Oracle reaches only 26\% on multi-scenario tasks.
This shortfall reflects a deficit in multi-step path formation across large topologies.
LLMs can reference named places and operate over simple pre-specified graphs, yet they rarely assemble the connectivity constraints that turn local doorways into a valid end-to-end route.
Rather than simply providing structured data and expecting a correct plan, a practical system must actively scaffold reasoning by constructing connectivity, verifying reachability, and validating long-horizon trajectories within the spatial graph.

\noindent\textbf{Gap 3: Optimization Shortfall in LLM Planning.} Beyond representation and reasoning, planning quality falters at the optimization level.
Even with perfect topological semantics, the Oracle still yields inefficient schedules, with overlap reaching 45\%.
Without structured guidance, the inefficiency compounds dramatically. On multi-scenario tasks, the Naive setting expends a staggering \(48\times\) more tokens than the Oracle.
These patterns clearly indicate LLMs inherently tend to satisfice rather than optimize, producing plausible yet resource-heavy plans~\cite{Wu2024CATPLLMEL,Liu2023ControlLLMAL}. Therefore, a practical system should capitalize on LLM’s semantic strengths while delegating global resource efficiency to deterministic algorithms\cite{Arora2024AnticipateA}.

\vspace{-0.16in}
\subsection{Motivation \& Core Ideas}
\label{subsec:motivation_paradigm}

The preceding findings yield a crucial insight that constructing a reliable and efficient semantic scheduling system cannot simply treat the LLM as an unconstrained, end-to-end black-box planner. These inherent, fundamental gaps in representation, reasoning, and optimization necessitate a novel paradigm to explicitly structure and guide the LLM's role.

This motivates our core idea to replace monolithic opaque planning with a structured and verifiable workflow centered on the \textit{Spatial Trajectory Graph (STG)}. Our neurosymbolic \textit{STG} paradigm decomposes the intractable scheduling problem into a principled three-stage process, each instantiated by a corresponding phase in our \textit{IoT-Brain} system. 1) \textit{Intent Formalization}~(\S\ref{subsec:semantic_structuring}) first leverages an LLM's semantic competence to translate a user's ambiguous query into a hypothesized \textit{STG}, a structured but unverified blueprint of intent. 2) \textit{Feasibility Grounding}~(\S\ref{subsec:symbolic_grounding}) then enters an iterative "verify-before-commit" loop where each element of the blueprint is systematically validated against the physical world model until a fully consistent and grounded \textit{STG} is produced. 3) \textit{Optimal Synthesis}~(\S\ref{subsec:synthesis_and_perception}) finally compiles the now-verified blueprint into a resource-optimal sensor activation plan and executes it adaptively using a perception-in-the-loop mechanism to handle unfolding real-world dynamics. The multi-stage process provides a systemically verifiable solution for LLM-based grounding, making a significant step towards enabling truly intelligent AIoT systems.

\vspace{-0.06in}
\section{S³ AND STG FORMULATION}
\label{sec:problem_and_paradigm}
We first formally define the \textit{Semantic–Spatial Sensor Scheduling (S³)} problem as a principled mapping from high-level user intent to a resource-aware dynamic activation plan~(\S\ref{subsec:problem_formulation}). Building on this formulation, we introduce the \textit{Spatial Trajectory Graph (STG)}, a paradigm that grounds free-form language into an optimization-ready spatial graph to systematically address this fundamental challenge~(\S\ref{subsec:stg_paradigm}).

\vspace{-0.1in}
\subsection{The S³ Problem}
\label{subsec:problem_formulation}

\noindent\textbf{Environment and Query.} We consider a large, densely instrumented site where a user issues a natural language query $Q_{NL}$. The environment is represented by a comprehensive world model \(W=(\mathcal{G}_{\mathrm{spatial}},\allowbreak \mathcal{G}_{\mathrm{sensor}},\allowbreak \Phi)\)
where \(\mathcal{G}_{\mathrm{spatial}}\) is a labeled spatial graph of locations and traversability,
\(\mathcal{G}_{\mathrm{sensor}}\) is a device graph of sensors and their capabilities,
and \(\Phi\) links the two by encoding sensor visibility and geometry.

\noindent\textbf{Semantic Compilation.}
Given \(Q_{\mathrm{NL}}\) and \(W\), semantic compilation yields a set of verifiable spatiotemporal predicates \(\Omega(Q_{\mathrm{NL}},\, W)\) specifying geographical anchors, spatial regions, relational constraints, and explicit temporal bounds (e.g., "11:30 a.m."). Let \(\Pi(\Omega)\) denote the admissible spatiotemporal witnesses, representing the trajectories that must be observed within a specific timeframe to satisfy the query.

\noindent\textbf{Plans and Objective.} Answering the query requires generating a dynamic activation plan $P(t)$, a time-varying set of sensors intended to reconstruct a witness trajectory $\tau \in \Pi(\Omega)$. A plan is feasible if its collective observation over time, denoted as $\mathcal{P} = \bigcup_{t} P(t)$, completely covers at least one witness $\tau$. The objective is to find an optimal plan $P^*(t)$ that maximizes a fidelity-cost trade-off, formalized as: 
\begin{equation}
\label{s3_problem}
P^*(t)\!=\!\underset{P(t)\subseteq \mathcal{G}_{\mathrm{sensor}}}{\arg\max}
\!\left(\!\mathcal{F}(\mathcal{P};\Pi(\Omega),\Phi)\!-\!\lambda\!\int\!\operatorname{Cost}(P(t)) dt\!\right)\!,
\end{equation}
where fidelity $\mathcal{F}$ measures the quality of the reconstructed trajectory, monotonically increasing with its completeness and accuracy. The $\operatorname{Cost}(P(t))$ function captures all resource expenditures, including the number of active sensors, activation duration, and redundant spatial overlap.

\noindent\textbf{Inherent Requirements.} While Eq.~\ref{s3_problem} specifies the optimization objective, solving it directly with off-the-shelf LLMs is intractable given the model’s intrinsic computational limits. Informed by our preliminary study~(\S\ref{subsec:preliminary_study}), a viable LLM-based approach must satisfy three fundamental requirements to align model capability with rigorous problem demands. 1) \textit{Semantic Grounding.} The solution must anchor the LLM's linguistic outputs to the physical world. Fuzzy descriptions like "near the main entrance" must be unambiguously resolved to concrete spatial entities in $W$ to become actionable. 2) \textit{Topological Feasibility.} The solution must enforce topological validity on the LLM's generated plans. Any long-horizon trajectory $\tau$ must be explicitly verified as physically traversable within $\mathcal{G}_{spatial}$, precluding the LLM from simply hallucinating impossible paths. 3) \textit{Resource Efficiency.} The solution must decouple planning from optimization. Given that LLMs are satisficers, not optimizers, a practical system must delegate the selection of a resource-efficient schedule to specialized deterministic algorithms. These requirements motivate a paradigm that structures and constrains the LLM’s role, moving beyond brittle end-to-end planning.

\begin{figure*}[t]
    \centering
    \setlength{\abovecaptionskip}{0.03cm}
    \includegraphics[width=\textwidth]{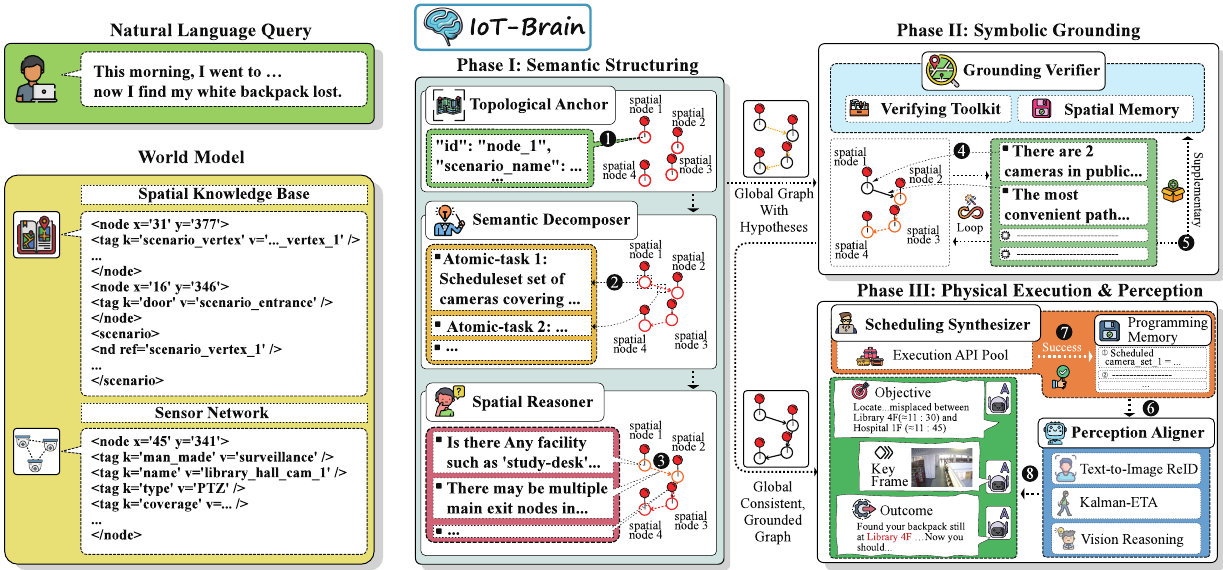}
    \caption{The system overview and workflow of \textit{IoT-Brain}.}
    \label{fig:system_architecture}
    \vspace{-4mm}
\end{figure*}

\vspace{-0.1in}
\subsection{The STG Paradigm}
\label{subsec:stg_paradigm}

End-to-end sensor selection couples semantics, topology, resources, and timing, hindering verification of intermediate assumptions. To make decisions verifiable and cost-aware, we introduce \textit{STG}. \textit{STG} decouples semantic interpretation, topological grounding, and resource optimization by inserting an explicit spatiotemporal trajectory graph where candidate hypotheses are checked before commitment. This blueprint transforms a user's intent into a structured trajectory, the trajectory into verified spatiotemporal facts, and these facts into an optimized, dynamic activation schedule.

\noindent\textbf{STG Definition.} An \textit{STG} is a quadruple $G=(V, E, \tau_V, \sigma_t)$ that provides a  verifiable specification for a dynamic plan. It comprises: (i) a connected subgraph \((V,E)\) of relevant locations; (ii) a verified spatial witness path \(\tau_V=(v_{1},\ldots,v_{m})\) that satisfies the spatial predicates; and (iii) a dynamic sensor scheduling function \(\sigma_t\), which maps each spatial node \(v_i \in \tau_V\) to a set of sensors to be activated at a dynamically determined time \(t_i\). The induced dynamic plan is thus $P(t) = \sigma_t(v_i)$ for $t=t_i$. An \textit{STG} is hypothesized (\(G_{0}\)) when its spatial path \(\tau_{V,0}\) is unverified, and becomes grounded (\(G_{\star}\)) only after $\tau_V$ is validated against the world model \(W\).

\noindent\textbf{Objective Reparameterization.} The \textit{STG} structure recasts the original optimization over dynamic plans $P(t)$ as a more tractable, two-stage process. First, it searches over the space of grounded spatial paths ($\mathcal{G}_{\star}$) to find an optimal $\tau_V^*$. Second, it determines the optimal temporal scheduling ($\sigma_t^*$) along that path. The objective is formalized as: 
\begin{equation}
\label{eq:stg_objective}
(\tau_V^*, \sigma_t^*)=\arg\max_{(\tau_V, \sigma_t)}
\Big(\mathcal{F}(\tau_V, \sigma_t)-\lambda\,\operatorname{Cost}(\tau_V, \sigma_t)\Big),
\end{equation}
where the optimal dynamic plan $P^*(t)$ is derived from $(\tau_V^*, \sigma_t^*)$. This reframing is pivotal because it separates the verifiable, static spatial planning from the adaptive, online temporal scheduling, making the problem tractable.

\noindent\textbf{Principled Inference Workflow.} The reframing supports a three-stage workflow guided by the "verify-before-commit" principle. i) \textit{Intent Formalization.} From the user query, construct an ungrounded graph \(G_{0}\) populated with candidate locations and a hypothesized spatial path \(\tau_{V,0}\). ii) \textit{Feasibility Grounding.} Enter an iterative loop that validates spatial hypotheses in \(\tau_{V,0}\) against the world model \(W\), disambiguates semantics, and enforces topological feasibility until a fully grounded spatial path \(G_{\star}\) emerges. iii) \textit{Optimal Synthesis.} On the grounded path \(G_{\star}\), compute the optimal dynamic scheduling function \(\sigma_t^{*}\) maximizes the objective in Eq.~\ref{eq:stg_objective}. This yields a verifiably resource-aware dynamic activation plan $P^*(t)$.
Within this workflow, the LLM proposes and refines spatial hypotheses, while the correctness of the spatial path and the optimality of the spatiotemporal schedule are secured by explicit checks and deterministic solvers.

\vspace{-0.1in}
\section{IOT-BRAIN SYSTEM ARCHITECTURE}
\label{sec:system_design}

To instantiate the \textit{STG} paradigm, we implemented \textit{IoT-Brain}, a modular framework that turns high-level semantic queries into verifiable, resource-aware sensor activation plans. As depicted in Fig.~\ref{fig:system_architecture}, \textit{IoT-Brain} follows a three-stage pipeline comprising Semantic Structuring (\S\ref{subsec:semantic_structuring}), Symbolic Grounding (\S\ref{subsec:symbolic_grounding}), and Adaptive Execution and Perception (\S\ref{subsec:synthesis_and_perception}).

\vspace{-0.1in}
\subsection{System Workflow}
\label{subsec:system_architecture}

\textit{IoT-Brain} takes two inputs, a natural language query \(Q_{\mathrm{NL}}\) and a world model \(W\) combining detailed spatial knowledge with a sensor-network map, and processes them through a structured and verifiable three-stage pipeline.

\noindent\textbf{Semantic Structuring.}
\rev{
The workflow employs three LLM agents as one-shot planners using system prompts.
First, the \textit{Topological Anchor} \ding{202} extracts geographical entities to map textual mentions in $Q_{\mathrm{NL}}$ to spatial graph nodes, seeding a scaffold.
Building on this, the \textit{Semantic Decomposer} \ding{203} factors the goal into a logical witness walk of atomic sub-tasks.
To bridge high-level plans and physical constraints, the \textit{Spatial Reasoner} \ding{204} analyzes transitions to formulate verifiable hypotheses regarding topological connectivity and attributes.
}

\noindent\textbf{Symbolic Grounding.}
\rev{
To validate these hypotheses, the \textit{Grounding Verifier} \ding{205} operates as an agent in an iterative Thought-Action-Observation loop~\cite{Yao2022ReActSR}.
It translates abstract hypotheses into concrete checks by invoking our custom-crafted deterministic \textit{Verifying Toolkit}, a Python library designed to query the explicit geometry and sensor coverage within \(W\).
This rigorous loop prunes topologically infeasible branches until a consistent grounded graph emerges.
Verified facts are cached as system-wide topological consensus in \textit{Spatial Memory} \ding{206} to accelerate future lookups.
}

\noindent\textbf{Adaptive Execution and Perception.}
\rev{
With the verified graph, the \textit{Scheduling Synthesizer} \ding{207} acts as a graph-to-code compiler, generating a Python script that invokes specialized heuristic solvers from the \textit{Execution API Pool} for resource-optimal scheduling.
Successful scripts are cached in \textit{Programming Memory} \ding{208} for efficient in-context reuse~\cite{Zhao2021CalibrateBU, Ma2023FairnessguidedFP}. 
The \textit{Perception Aligner} \ding{209} then executes this schedule through a seamlessly coordinated pipeline: a VLM first grounds the text description to a visual target, passing the visual embedding to a Re-ID network for cross-camera association, while a Kalman-ETA filter~\cite{Achar2020BusAT} predicts arrival times to trigger downstream sensors just-in-time precisely.  
}

\vspace{-0.16in}
\subsection{Intent-to-Blueprint Structuring}
\label{subsec:semantic_structuring}

\begin{figure}[t]
    \centering
    \setlength{\abovecaptionskip}{0.cm}
    \includegraphics[width=\columnwidth]{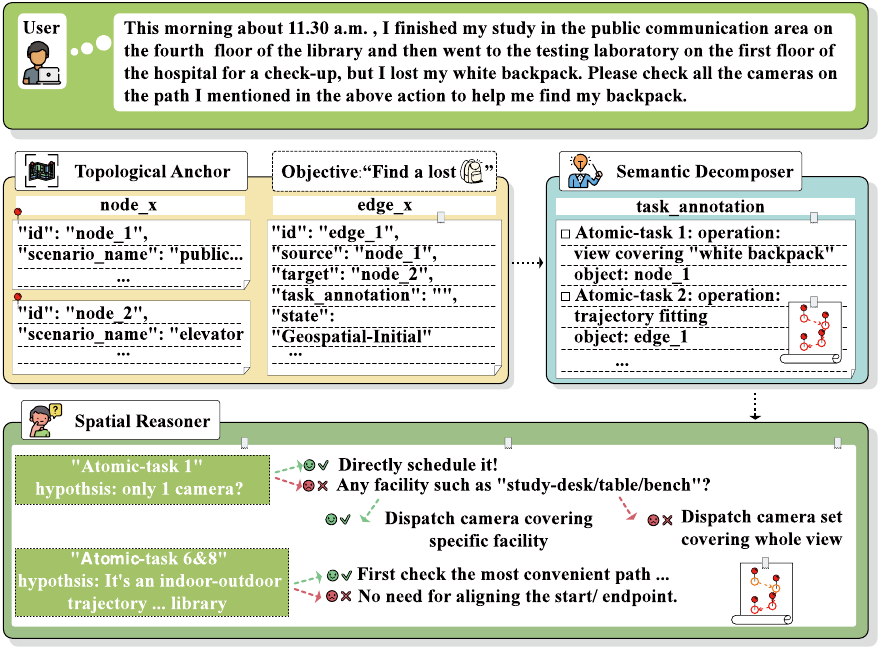}
    \caption{Dataflow of the Semantic Structuring phase.}
    \label{fig:dataflow_structuring}
    \vspace{-5mm}
\end{figure}

\rev{
The initial phase of \textit{IoT-Brain} converts the unstructured user query \(Q_{\mathrm{NL}}\) into a structured representation of intent and context. This process, termed \textit{Semantic Structuring}, progressively builds a hypothesized \textit{STG}, which is a machine-interpretable blueprint derived from the query but not yet verified against the world model. Fig.~\ref{fig:dataflow_structuring} depicts the three-agent pipeline, illustrated using the "lost backpack" query.
}

\begin{figure*}[t]
    \centering
    \setlength{\abovecaptionskip}{0.03cm}
    \includegraphics[width=1\textwidth]{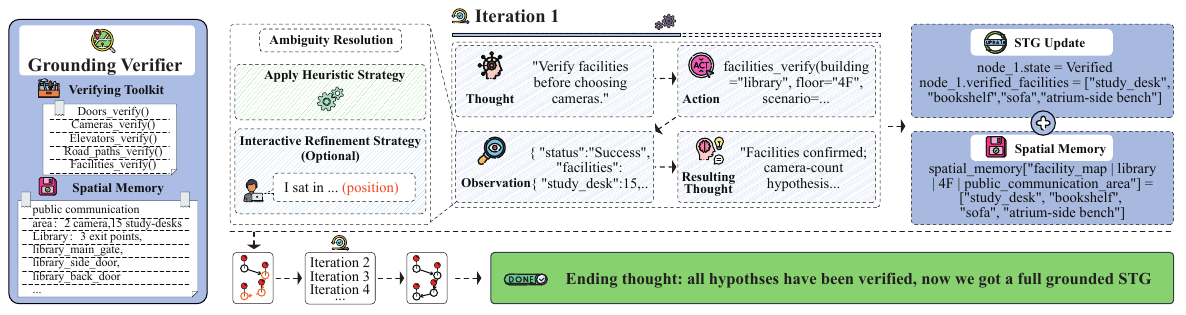}
    \caption{Workflow of the Hypothesis-Verification Loop.}
    \label{fig:workflow_grounding}
    \vspace{-4mm}
\end{figure*}
\noindent\textbf{\ding{202} Topological Anchor.}
\rev{
The pipeline begins with the \textit{Topological Anchor}, a coarse-grained parser that identifies the spatial scope of the user's intent. 
By leveraging topological priors, the agent instantiates a candidate subgraph containing both explicitly mentioned locations and implicitly required transition points.
For the running example, the agent identifies the defined starting \texttt{public communication area} and the target destination \texttt{testing laboratory}, while simultaneously deducing necessary intermediate connectors, such as the \texttt{Library elevator} (shown as \texttt{node\_2} in Fig.~\ref{fig:dataflow_structuring}), required to bridge the vertical floor transition.
These explicit and implicit geographical entities collectively form the initial, unverified vertices \(V_0\) and edges \(E_0\) of a nascent \textit{STG}, effectively representing the initial structural hypothesis of the user's underlying intended spatial context.
}

\noindent\textbf{\ding{203} Semantic Decomposer.}
\rev{
While the \textit{Anchor} provides disconnected spatial candidates, the \textit{Semantic Decomposer} imposes order by factoring high-level intent into a sequence of atomic operations mapped to these nodes.
This step ensures topological coherence by constructing a valid traversal that chains these entities, such as planning a continuous route from the starting \texttt{Library 4F} through inferred connectors like the \texttt{Elevator} to the \texttt{Hospital 1F}. This process defines the spatial witness walk $\tau_{V,0}=(v_1, \ldots, v_m)$, enriching the graph with a hypothesized trajectory and task annotations. 
}

\noindent\textbf{\ding{204} Spatial Reasoner.}
\rev{
However, this planned trajectory remains speculative as it relies on high-level topological priors rather than physically grounded facts. 
Implicit assumptions arise wherever the semantic logic of the planner might diverge from strict physical availability (e.g., assuming a specific door is unlocked or a pathway is currently traversable).
The \textit{Spatial Reasoner} systematically bridges this gap by scrutinizing the entire generated trajectory to convert these implicit assumptions into explicit, deterministic, and verifiable hypotheses.
In the context of the library-to-hospital transition, it detects potential ambiguity and hypothesizes that a specific, valid exit must be confirmed against the world model. Similarly, for the area covering task, it formulates hypotheses regarding the existence of specific facilities (e.g., "study desks") to  narrow the intended sensing scope. 
}

\rev{
Each hypothesis is cast as a concrete, verifiable proposition regarding the world model, defined with an explicit scope and a wide range of diverse admissible evidence sources. 
This collaborative three-agent pipeline collectively returns a comprehensive hypothesized spatial path, encapsulated in the initial \textit{STG}, \(G_{0}=(V_{0}, E_{0}, \tau_{V,0}, \emptyset)\), that is now fully ready for the critical and rigorous \textit{Symbolic Grounding} phase. 
}

\vspace{-0.12in}
\subsection{Hypothesis-to-Fact Grounding}
\label{subsec:symbolic_grounding}

\rev{
A detailed hypothesized graph remains non-actionable as long as its elements are unverified.
The \textit{Symbolic Grounding} phase converts the speculative \textit{STG}, \(G_{0}\) into a grounded \textit{STG}, \(G_{\star}\) by rigorously testing each hypothesis against the world model \(W\).
As shown in Fig.~\ref{fig:workflow_grounding}, a "verify-before-commit" loop promotes facts, prunes contradictions, and resolves ambiguities until the graph is verifiably consistent with \(W\).
}

\noindent\textbf{\ding{205} The Hypothesis-Verification Loop.} 
\rev{
Grounding hinges on a rigorous verification loop executed by the \textit{Grounding Verifier}, operating as a tool-using agent capable of query-based interaction with the physical world model. 
Its primary responsibility is to eliminate ambiguity by converting semantic assumptions into verifiable topological facts. 
Through a Thought-Action-Observation cycle~\cite{Yao2022ReActSR, Shinn2023ReflexionLA}, the agent addresses specific node hypotheses, such as identifying a valid exit in the \texttt{Library}, translating them into designing a corresponding topological verification query. 
It executes this design via invoking the crafted \textit{Verifying Toolkit} (e.g., \texttt{doors\_verify()}) to retrieve concrete spatial data from the environment. 
The resulting observation serves as ground truth, enabling the agent to define the precise sensor scheduling logic for that location and progressively transform the speculative skeleton into a fully grounded \textit{STG}.
}

\noindent\textbf{Resolving Ambiguity.} 
\rev{
A critical challenge emerges when the retrieved observation is not definitive, such as the toolkit returning multiple valid exits for the \texttt{Library}. 
To resolve the uncertainty deterministically, \textit{IoT-Brain} employs a recursive reasoning strategy based on topological consistency. 
Upon detecting multiple potential exits, the agent initiates a secondary check to evaluate the connectivity of each candidate relative to the subsequent waypoint (the \texttt{Hospital}). 
By filtering out exits that lack a valid traversable path to the destination, the system autonomously identifies the topologically sound option. 
This heuristic ensures the final plan is physically executable without human intervention, though an optional interactive strategy is available for extreme cases where user clarification is preferred.
}

\noindent\textbf{\ding{206} Amortized Verification via Caching.} 
\rev{
Throughout the iterative process, successfully verified facts are cached according to their associated locations in a \textit{Spatial Memory} module, a technique inspired by classic indexing and memoization~\cite{Lewis2020RetrievalAugmentedGF}. 
This mechanism effectively amortizes the computational overhead of repeated queries for the same entities (e.g., retaining the validated \texttt{Library} exit details for future requests), thereby significantly accelerating future verification. 
The verification loop terminates once all the structural hypotheses in \(G_{0}\) are resolved. 
The final output \(G_{\star}\) represents a verified spatial blueprint where every node, edge, and the contained spatial path \(\tau_V\) are consistent with the physical world, thereby providing a solid and reliable foundation for the subsequent scheduling optimization. 
}

\begin{figure}[t]
    \centering
    \setlength{\abovecaptionskip}{0.03cm}
    \includegraphics[width=\columnwidth]{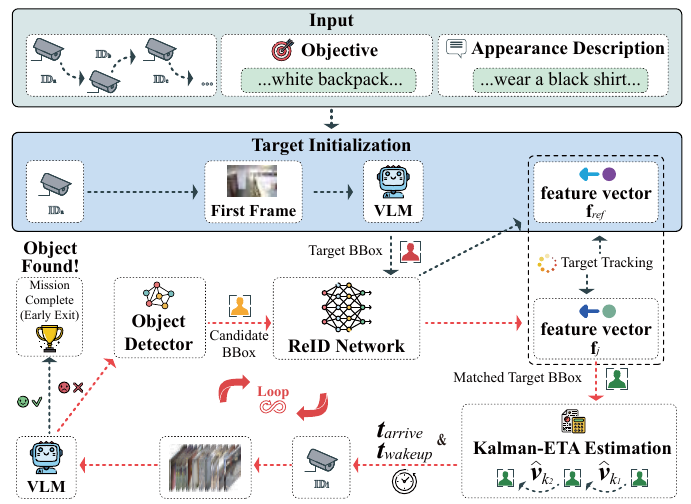}
    \caption{Workflow of the \textit{Perception Aligner}.}
    \label{fig:perception_aligner}
    \vspace{-2mm}
\end{figure}

\begin{figure*}[t]
    \centering
    \setlength{\abovecaptionskip}{0.03cm}
    \includegraphics[width=\textwidth]{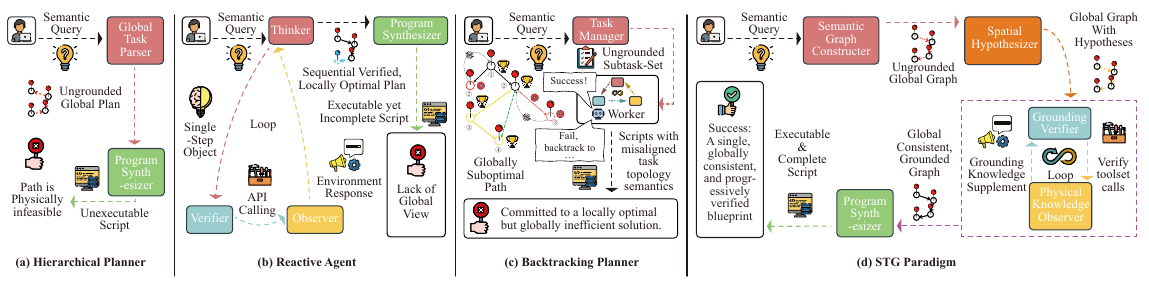} 
    \caption{Conceptual workflow comparison of agentic paradigms. We contrast the brittle Hierarchical approach, the myopic Reactive agent, and the costly Backtracking planner with our structuring verifiable \textit{STG} paradigm.}
    \label{fig:paradigm_comparison}
    \vspace{-5mm}
\end{figure*}

\vspace{-0.12in}
\subsection{From Plan to Optimized Action}
\label{subsec:synthesis_and_perception}

With a fully grounded spatial blueprint \(G_{\star}\) in place, the final phase of \textit{IoT-Brain} translates it into dynamic action and perception in the physical world. The workflow comprises two components to achieve resource optimization synergistically.
The first is a deterministic compiler that synthesizes an executable and resource-aware plan.
The second is an adaptive executor that manages real-time operation.

\noindent\textbf{\ding{207} Scheduling Synthesizer.}
\rev{
With the verified spatial blueprint \(G_{\star}\) established, the \textit{Scheduling Synthesizer} translates the plan into optimized action.
Functioning as a graph-to-code compiler, it transforms the grounded nodes and edges into an executable Python script.
This stage operationalizes the resource optimization objective in Eq.~\ref{eq:stg_objective} by mapping the verified sub-tasks to specialized functions within our \textit{Execution API Pool}.
Consequently, the validated path segment inside the \texttt{Library} is processed by generating a call to \texttt{indoor\_path\_camera\_search()}, which utilizes a deterministic set-cover algorithm~\cite{Zhu2022OptimizingND} to select the minimal sensor set required for full visibility.
By embedding these solvers within a deterministic API layer, the compiler ensures the final schedule strictly honors the topological constraints verified in the previous phase while maximizing resource efficiency. 
Successful query--script pairs are cached in \textit{Programming Memory} \ding{208} to accelerate future compilations on semantically-related tasks.
}

\noindent\textbf{\ding{209} Perception Aligner.} 
\rev{
A static script is insufficient for dynamic, real-world execution. 
The \textit{Perception Aligner} therefore operates as an online executor, tightly coupling perception with control to keep only necessary sensors active. 
As depicted in Fig.~\ref{fig:perception_aligner}, the process commences with target grounding, where a VLM anchors the user's textual description of the "\texttt{white backpack}" to a specific visual instance in an initial video frame~\cite{Tan2024HarnessingTP,Jiang2025ModelingTO}. 
To solve the association problem of maintaining the target's identity across a distributed camera network, we employ a robust feature extractor (e.g., a Re-ID network) to match visual appearances~\cite{Hu2024PersonViTLS}. 
Concurrently, a predictive model (e.g., a Kalman filter) estimates arrival times at downstream viewpoints along the planned route, enabling just-in-time sensor activation to optimize resource usage~\cite{Yu2019ActivityNetQAAD,Yang2022LearningTA,Basagni2019WakeupRR}. 
Finally, the same VLM performs continuous verification, conducting frame-level reasoning to evaluate task predicates. 
This allows the system to intelligently decide when the query is satisfied, triggering early termination to conserve resources~\cite{Agrawal2015VQAVQ}. 
This unified loop reduces overhead by activating sensors only when needed and deactivating them once sufficient proof is obtained.
}

\vspace{-0.1in}
\section{EVALUATION ON TOPOSENSE-BENCH}
\label{sec:evaluation_benchmark}

We conduct a comprehensive evaluation on our large-scale benchmark, \textit{TopoSense-Bench}, to empirically validate the \textit{STG} paradigm and quantify the performance of \textit{IoT-Brain}. Our highlights are as follows:
\begin{itemize}[leftmargin=*, topsep=0pt, partopsep=0pt, itemsep=0pt, parsep=0pt]
    \item \textit{IoT-Brain} delivers superior reliability and efficiency. On complex tasks, it boosts task success by up to 7.4$\times$ over the classical Hierarchical planner\cite{Ge2023OpenAGIWL,Shen2023HuggingGPTSA} and runs nearly 2$\times$ faster with 6.6$\times$ fewer prompt tokens than the search-intensive Backtracking planner\cite{Qin2023ToolLLMFL,Chen2024SmurfsMS} (\S\ref{subsec:e2e_performance}).
    \item \textit{IoT-Brain} scales and generalizes well. Its verification overhead grows near-linearly with query complexity, not exponentially. Furthermore, its performance gains persist across diverse foundation models, confirming the benefits stem from our architecture, not a specific backbone (\S\ref{subsec:sensitivity}).
    \item \textit{IoT-Brain} derives strength from the full \textit{STG} pipeline. Ablations confirm the synergy of its core components and reveal that unverified hypotheses are actively harmful, reinforcing "verify-before-commit" as a prerequisite for reliable physical-world planning (\S\ref{subsec:ablation}).
\end{itemize}

\vspace{-0.1in}
\subsection{Experimental Setup}
\label{subsec:setup}

\noindent\textbf{The TopoSense-Bench Benchmark.} 
\rev{
To rigorously evaluate our systems, we constructed \textit{TopoSense-Bench}, a large-scale benchmark for the \textit{S\textsuperscript{3}} problem. 
Its central design principle is semantic textualization, transforming raw OSM data into a structured knowledge base. 
We normalize ontological tags~\cite{osm_taginfo}, generate hierarchical human-readable names~\cite{nominatim_manual,lawrence_nlmaps_2016}, and project geodetic coordinates into a site-local Cartesian frame, creating a unified substrate where physical topology and sensor capabilities are jointly considered.
The benchmark instantiates a realistic campus spanning 33 buildings, 161 floor plans, and a dense network of 2,510 cameras.}

\rev{
Layered on this environment is a suite of 5,250 natural language queries grounded in the operational realities of large-scale sensor networks. 
Since providing spatiotemporal anchors is a realistic prerequisite for initiating tasks, the core \(S^3\) challenge focuses on reasoning over the extensive topological knowledge base to infer the complete, optimal sensor path between these points.
To ensure consistently high data quality, we employed a rigorous three-stage construction pipeline. 
Domain experts first authored $\sim$200 base templates per tier to cover a wide spectrum of realistic scenarios. 
Using these as seeds, GPT-o3~\cite{openai2025o3systemcard} synthesized thousands of distinct queries by instantiating diverse semantic intents across the complex campus topology. 
Crucially, every resulting query underwent comprehensive, rigorous manual expert verification and ground-truth annotation to guarantee logical soundness and topological solvability, which ultimately ensures the benchmark's high  fidelity. 
}

\begin{table}[t]
\centering
\setlength{\abovecaptionskip}{0.03cm}
\caption{Statistics \& Taxonomy of \textit{TopoSense-Bench}.}
\label{tab:benchmark_stats}
\resizebox{\columnwidth}{!}{%
\begin{tabular}{@{}lllr@{}}
\toprule
\multicolumn{2}{l}{\textbf{Knowledge Base Statistics}} & \multicolumn{2}{r}{\textbf{Value}} \\
\midrule
\multicolumn{2}{l}{Buildings / Floor Plans / Outdoor Segments} & \multicolumn{2}{r}{33 / 161 / 53} \\
\multicolumn{2}{l}{Total Topological Scenarios / Deployed Cameras} & \multicolumn{2}{r}{7,832 / 2,510} \\
\midrule
\multicolumn{2}{l}{\textbf{Query Dataset Statistics}} & \multicolumn{2}{r}{\textbf{Count (\%)}} \\
\midrule
\multicolumn{4}{l}{\textit{\textbf{Tier 1: Intra-Zone Perception}}} \\
& \multicolumn{2}{l}{T1.a: Focal Scene Awareness} & 1,433 (27.3\%) \\
& \multicolumn{3}{l}{\textit{\quad e.g., "verify activity near the door to room 5F 1"}} \\
& \multicolumn{2}{l}{T1.b: Panoramic Coverage} & 1,129 (21.5\%) \\
& \multicolumn{3}{l}{\textit{\quad e.g., "how many people are in the conference hall?"}} \\
\addlinespace
\multicolumn{4}{l}{\textit{\textbf{Tier 2: Intra-Building Coordination}}} \\
& \multicolumn{2}{l}{T2: Intra-Building Coordination} & 988 (18.8\%) \\
& \multicolumn{3}{l}{\textit{\quad e.g., "I lost a notebook between lab-8 and lab-7"}} \\
\addlinespace
\multicolumn{4}{l}{\textit{\textbf{Tier 3: Inter-Building Coordination}}} \\
& \multicolumn{2}{l}{T3.a: Open-Space Coordination} & 946 (18.0\%) \\
& \multicolumn{3}{l}{\textit{\quad e.g., "track my path from the west gate to the tennis court"}} \\
& \multicolumn{2}{l}{T3.b: Hybrid Indoor-Outdoor} & 754 (14.4\%) \\
& \multicolumn{3}{l}{\textit{\quad e.g., "I walked from building-1 to building-2..."}} \\
\bottomrule
\end{tabular}%
}
\vspace{-3mm}
\end{table}

\begin{figure*}[t]
    \centering
    \setlength{\abovecaptionskip}{0.03cm}
    \includegraphics[width=\textwidth]{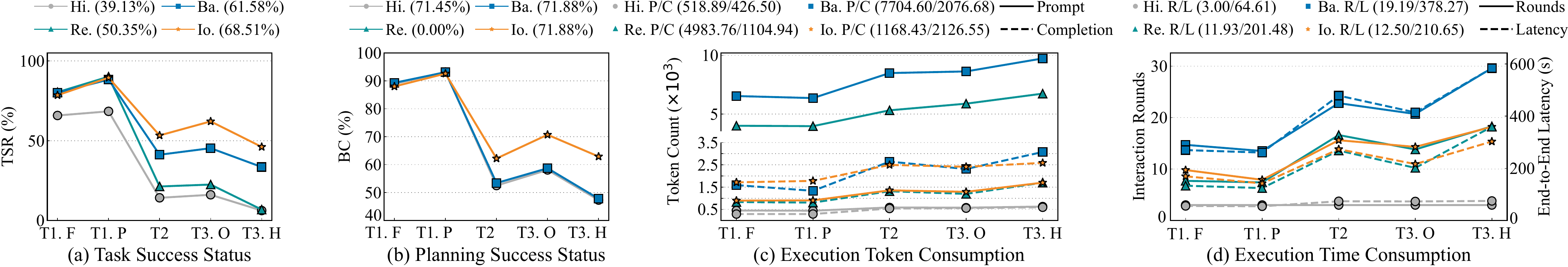}
    \caption{The overall performance on \textit{TopoSense-Bench}. We evaluate Hierarchical (Hi.), Reactive (Re.), Backtracking (Ba.), and \textit{IoT-Brain} (Io.) across five task categories: \rev{T1.F (Focal Scene), T1.P (Panoramic), T2 (Intra-Building), T3.O (Open-Space), and T3.H (Hybrid)}. Metrics include task success rate, blueprint correctness, token usage, iteration rounds, and end-to-end latency. The legend displays the average performance for each paradigm.}
    \label{fig:main_performance}
    \vspace{-5mm}
\end{figure*}

\noindent\textbf{Compared Paradigms.} We benchmark \textit{IoT-Brain} against agentized implementations of three influential LLM planning paradigms. To ensure comparability, all systems use the same foundation LLMs and API toolkits, differing only in their reasoning and execution policies.
(1) The Hierarchical planner~\cite{Ge2023OpenAGIWL,Shen2023HuggingGPTSA} follows a decompose-then-execute strategy. It produces a complete, high-level plan upfront without intermediate verification, yielding speed at the cost of brittleness.
(2) The Reactive planner~\cite{Yao2022ReActSR,Chen2024ActiveIF} instantiates the Thought-Action-Observation loop. It operates step-by-step, making it highly adaptive but often myopic on long-horizon tasks.
(3) The Backtracking planner~\cite{Qin2023ToolLLMFL,Chen2024SmurfsMS} is inspired by depth-first search over a decision tree. It improves reliability by exploring multiple branches, but this expanded search typically incurs prohibitive inference costs and can commit to locally optimal yet globally inefficient paths.
Fig.~\ref{fig:paradigm_comparison} contrasts these workflows with our \textit{STG} approach.

\noindent\textbf{Evaluation Metrics.} We assess each paradigm's performance from the dual perspectives of reliability and efficiency. We use a suite of standardized metrics:
For reliability, we measure (1) task success rate (TSR), the percentage of queries yielding a functionally correct final answer, and (2) blueprint correctness (BC), which evaluates if a pre-execution plan is logically sound and topologically feasible, scored via an LLM-as-a-Judge protocol~\cite{Zheng2023JudgingLW}.
For efficiency, we quantify (3) inference cost, measured in total LLM tokens; (4) interaction rounds, the number of agent–LLM turns; and (5) end-to-end latency, the total time from query submission to final answer.

\vspace{-0.12in}
\subsection{End-to-End Performance}
\label{subsec:e2e_performance}

\noindent\textbf{Reliability Analysis.} Fig.~\ref{fig:main_performance}(a-b) presents the TSR and BC across all task categories and difficulty tiers. A key observation is that while baseline planners often struggle to translate symbolically correct plans into real-world success, \textit{IoT-Brain} consistently maintains high TSR, especially as task complexity increases to long-horizon settings. This empirically demonstrates the \textit{STG} paradigm's effectiveness in resolving the fundamental gaps that plague end-to-end LLM planning.

A notable paradox emerges on simple Tier 1 tasks. The plan-less Reactive planner achieves a strong TSR, demonstrating for simple tasks, adaptive tool-use can suffice in easy settings. The Hierarchical planner, by contrast, highlights the critical gap between symbolic planning and physical execution. It frequently produces plausible blueprints with high BC scores, yet achieves markedly lower TSR. This discrepancy reflects the representation and reasoning gaps, where a plan is logically sound on paper proves physically unrealizable without grounding in the real world. This confirms a correct blueprint is necessary but not sufficient for success.

As complexity escalates to Tier 2 and 3, the necessity of structured grounding becomes indisputable. The performance of all baseline planners degrades sharply, which is a clear manifestation of the reasoning gap. Their inability to construct coherent, long-horizon plans leads to systemic failure. In stark contrast, on the most complex T3.Hybrid task, IoT-Brain achieves a 46.1\% TSR. This not only represents a more than 7.4$\times$ improvement over the Hierarchical planner's 6.2\% but also surpasses the strongest search-intensive alternative, the Backtracking planner, by 37.6\%. This sustained performance is a direct result of \textit{STG}'s "verify-before-commit" process, which constructs a globally coherent blueprint and systematically closes the reasoning gap.

\noindent\textbf{Efficiency Analysis.} Reliability must also be delivered efficiently. The computational overhead for each framework is reported in Fig.~\ref{fig:main_performance}(c–d) show that the cost escalates with task complexity for unstructured planners, exposing an optimization gap. The Hierarchical planner is fast yet unreliable, while Reactive and Backtracking exhibit rapidly increasing token consumption and latency as complexity grows. In pursuit of reliability through exhaustive search, Backtracking reaches nearly 600\,s on the most complex Tier 3 task.

\textit{IoT-Brain} closes this gap by casting planning as constrained optimization on a verified graph, yielding a markedly different efficiency profile. Its resource use grows more smoothly with complexity. The principled verification loop is more focused than the brute-force exploration of Backtracking and less myopic than the trial-and-error of Reactive. While delivering superior reliability, \textit{IoT-Brain} is nearly 2$\times$ faster on the most complex tasks and uses, on average, 6.6$\times$ fewer prompt tokens than Backtracking. These results show that the \textit{STG} paradigm not only improves reliability but also provides a scalable path to computational efficiency, achieving a win-win in both correctness and cost.

\begin{figure*}[t]
    \centering
    \setlength{\abovecaptionskip}{0.03cm}
    \includegraphics[width=\textwidth]{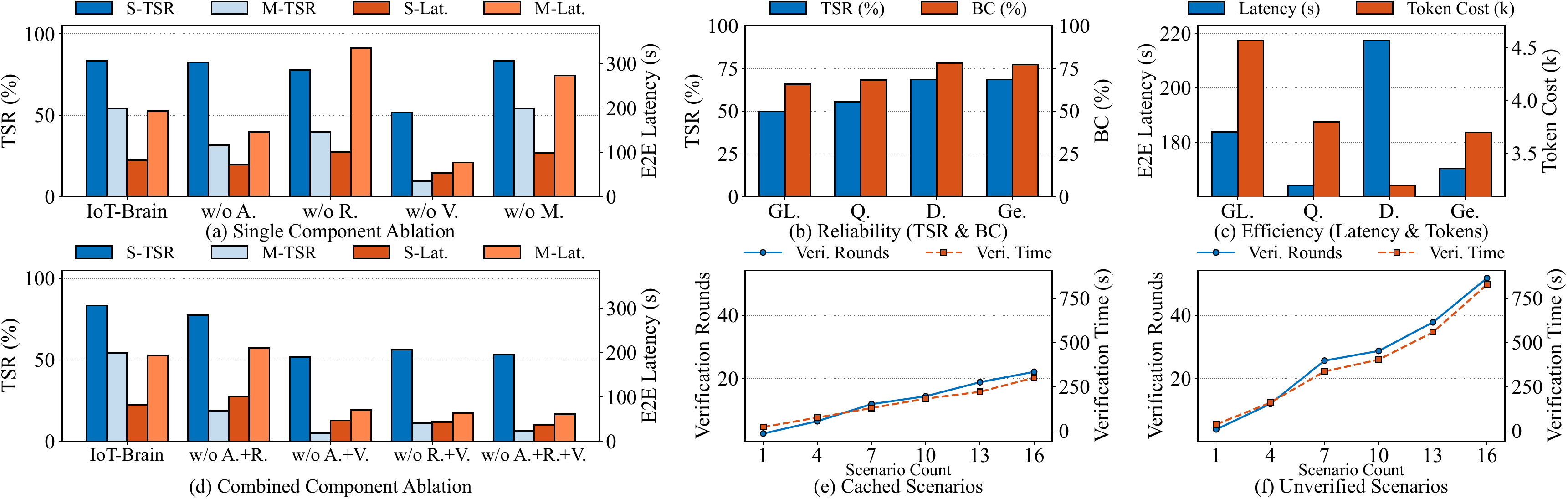}
    \caption{(a) \& (d): Ablation study of \textit{IoT-Brain}'s core components on single and multi scenario tasks. (A.: Anchor, R.: Reasoner, V.: Verifier, M.: Memory, S: Single-Scenario, M: Multi-Scenario). (b) \& (c): Sensitivity to different LLMs. (GL.: GLM-4, Q.: Qwen-Max, D.: DeepSeek-V3, Ge.: Gemini-2.5-Flash). (e) \& (f): Scalability with query complexity.  }
    \label{fig:sensitivity_llm}
    \vspace{-5mm}
\end{figure*}

\vspace{-0.12in}
\subsection{Scalability and Sensitivity Analysis}
\label{subsec:sensitivity}

\noindent\textbf{Sensitivity to Foundation Models.} To demonstrate that the benefits of \textit{STG} are not tied to a single proprietary model, we evaluate \textit{IoT-Brain}'s performance across four foundation LLMs. As expected, more powerful models yield higher end-to-end reliability (Fig.~\ref{fig:sensitivity_llm}(b)), confirming that the quality of the underlying LLM is a significant factor. More importantly, the efficiency profile shows no orders-of-magnitude variation in latency and token cost across models (Fig.~\ref{fig:sensitivity_llm}(c)). While different APIs exhibit distinct latency characteristics, the token consumption remains relatively stable. These results indicate \textit{STG} supplies a strong scaffold that channels reasoning into verifiable structure, allowing less capable models to perform competitively and confirming that observed gains arise from our architecture, not from any specific LLM's capabilities.

\noindent\textbf{Scalability with Query Complexity.} We investigate how \textit{IoT-Brain}'s computational verification overhead scales with query complexity, a key determinant of real-world usability. For cached scenarios, verification overhead grows sublinearly, as the system intelligently reuses entries from \textit{Spatial Memory} to amortize costs (Fig.~\ref{fig:sensitivity_llm}(e)). \rev{In stark contrast, for novel, unverified scenarios, the number of verification rounds increases in a near-linear fashion purely dependent on the count of distinct locations referenced in the query, rather than the global environment size} (Fig.~\ref{fig:sensitivity_llm}(f)). This predictable and bounded growth stands in sharp contrast to the exponential combinatorial blowup typical of unconstrained planning and demonstrates the inherent scalability of our structured, hypothesis-driven verification process.

\vspace{-0.12in}
\subsection{Ablation Study}
\label{subsec:ablation}

\noindent\textbf{Impact of Individual Components.} Fig.~\ref{fig:sensitivity_llm}(a) quantitatively shows the indispensable role of each component. Notably, removing the \textit{Grounding Verifier} is catastrophic, causing the multi-scenario TSR to plummet below 10\%, confirming the necessity of a "verify-before-commit" loop to prevent hallucination-prone planning. Removing the \textit{Spatial Reasoner} also severely degrades performance, especially on complex tasks, as the \textit{Verifier} is consequently forced into computationally expensive exhaustive checks without the \textit{Reasoner}'s focused hypotheses. Similarly, omitting the \textit{Topological Anchor} nearly halves multi-scenario TSR, underscoring its importance in providing the initial scaffolding for long-horizon plans. Finally, eliminating the \textit{Memory} modules, while not impacting single-run TSR, nearly doubles latency in complex settings, demonstrating their critical value in efficiently amortizing repetitive verification costs.

\noindent\textbf{Synergy of Combined Components.} Fig.~\ref{fig:sensitivity_llm}(d) underscores the components' tight complementarity. A variant lacking all core modules performs as poorly as the Hierarchical baseline, with its multi-scenario TSR collapsing to 6.4\%, confirming that the full pipeline is indispensable for robust behavior. More tellingly, a variant that isolates the \textit{Spatial Reasoner} by removing its structuring and verification counterparts performs even worse, with its multi-scenario TSR dropping to a mere 5.1\%. This result reveals a crucial, counter-intuitive insight that without validity checks, the \textit{Reasoner}'s unverified hypotheses are not merely neutral but actively harmful, introducing strong, misleading biases that derail the entire planning process. Collectively, these results powerfully validate our philosophy that a structured, verifiable grounding process is not an add-on, but the fundamental prerequisite for safe and reliable physical-world planning.

\vspace{-0.12in}
\section{EVALUATION ON A PHYSICAL TESTBED}
\label{sec:evaluation_real_world}

We complement our benchmark results with an end-to-end evaluation of \textit{IoT-Brain} in a large-scale, physical testbed. The experiments confirm the \textit{STG} paradigm's effectiveness in a real-world deployment, yielding a near-optimal balance of reliability and resource efficiency. Our highlights are:
\begin{itemize}[leftmargin=*, topsep=0pt, partopsep=0pt, itemsep=0pt, parsep=0pt]
    \item \textit{IoT-Brain} demonstrates strong resource efficiency, achieving a 49.84\% TCR that approaches the reliability upper bound while using \(4.1\times\) less bandwidth and processing \(4.2\times\) fewer frames than the resource-agnostic method (\S\ref{subsec:e2e_performance_real}).
    \item \textit{IoT-Brain}'s architecture is both efficient and robust. Latency profiling confirms its planning engine is swift, with dominant costs arising from task-intrinsic complexity. Being model-agnostic, the framework allows practitioners to flexibly balance performance, cost, and privacy by integrating diverse VLM backbones (\S\ref{subsec:sensitivity_real}).
\end{itemize}

\vspace{-0.12in}
\subsection{Testbed Configuration}
\label{subsec:testbed_config}

\begin{figure}[t]
    \centering
    \setlength{\abovecaptionskip}{0.1cm}
    \includegraphics[width=\columnwidth]{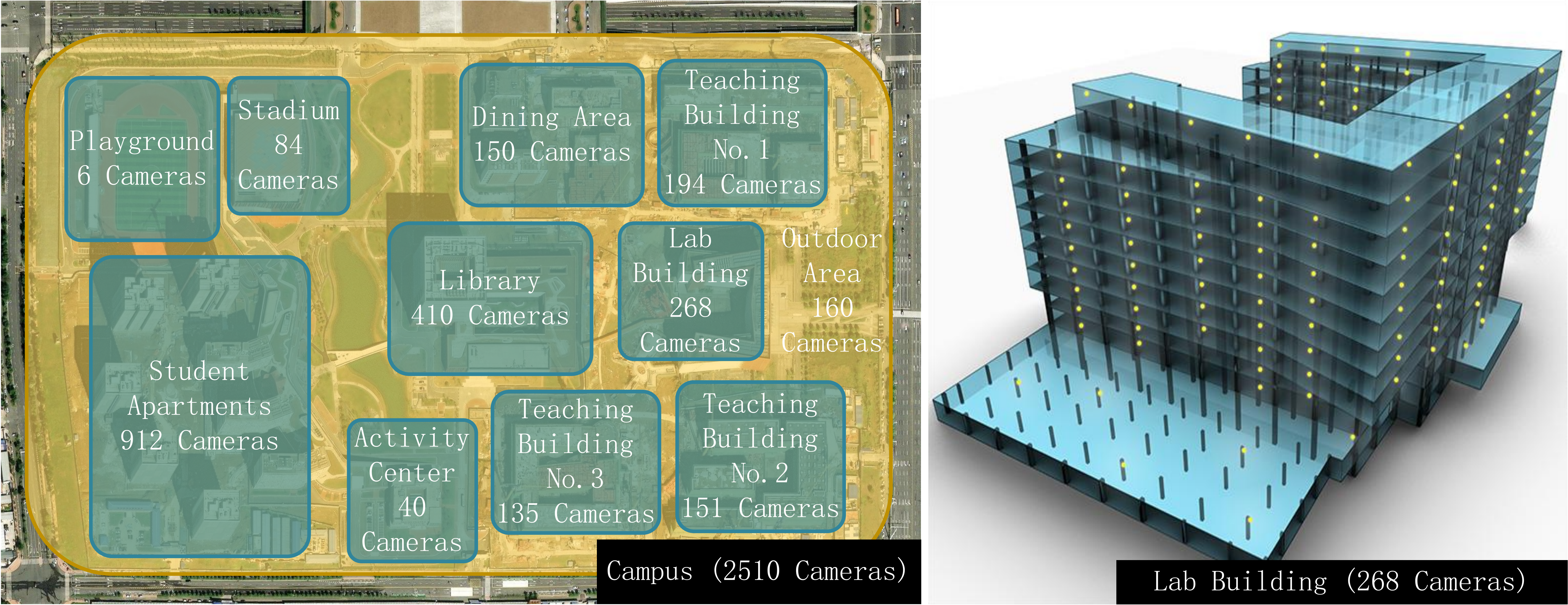}
    \caption{The real-world testbed environment.}
    \label{fig:testbed_map}
    \vspace{-2mm}
\end{figure}

\noindent\textbf{Physical Environment.} Our real-world testbed is a large-scale university campus instrumented with 2{,}510 Hikvision IP cameras distributed across 11 major areas. This diverse environment presents significant heterogeneity, where coverage ranges from sparse outdoor road networks to dense indoor deployments, with the Lab Building alone hosting 268 cameras covering complex topologies (see Fig.~\ref{fig:testbed_map}).

\noindent\textbf{System Deployment.} The \textit{IoT-Brain}\footnote{Project page: \url{https://github.com/houqiii/IoT-Brain}} system runs on a centralized server equipped with two NVIDIA A100 GPUs. Its core planning engine utilizes the Gemini-2.5-Flash API\cite{google_gemini25_flash_card} for efficient reasoning, while the \textit{Perception Aligner} executes on-premises using a compact and efficient visual stack that includes YOLOv8\cite{ultralytics_yolov8} for real-time detection and PersonViT-B/16\cite{hustvl_personvit_github} for robust image re-identification. All components communicate over the standard campus Wi-Fi infrastructure, reflecting a practical deployment setting constrained by realistic bandwidth fluctuations.

\vspace{-0.12in}
\subsection{Real-World Deployment Results}
\label{subsec:e2e_performance_real}

\noindent\textbf{Evaluation Protocol.} We evaluated end-to-end performance on 587 annotated real-world trajectories across three paradigms. In addition to our \textit{IoT-Brain} framework, we considered two baselines to benchmark against both common practice and a theoretical optimum: (1) Static Scheduling, which mirrors conventional security practice by triggering downstream sensors using a constant-velocity pedestrian model\cite{Schller2019TheST}, and (2) Naive Parallel Scheduling, a resource-agnostic upper bound that activates all potentially relevant cameras simultaneously. To ensure a controlled comparison, all vision–language queries were handled by a locally deployed Qwen-VL-Chat model\cite{Bai2023QwenVLAF}. Performance was measured using a comprehensive suite of metrics, including task completion rate (TCR), end-to-end latency, network bandwidth, and total frames processed (TFP).

\noindent\textbf{Performance Analysis.} Tab.~\ref{tab:paradigm_comparison_real} highlights the trade-off between reliability and resource use. Static Scheduling is frugal yet brittle, reaching only 3.61\% TCR and failing systematically whenever a target deviates from its pre-defined coverage. At the other extreme, Naive Parallel establishes an empirical upper bound on reliability at 65.64\% TCR, but does so at untenable cost, consuming $4.1\times$ more bandwidth and processing $4.2\times$ more frames than our system.

\textit{IoT-Brain} strikes an optimal balance within this trade-off.
Driven by the \textit{STG} paradigm for intelligent planning and the \textit{Perception Aligner} for just-in-time execution, it achieves a high TCR of 49.84\%, approaching the empirical upper bound, while keeping a resource footprint comparable to the far less reliable static strategy that uses the least bandwidth and processing the fewest frames among all paradigms.
Its sequential, plan-informed activation prevents the heavy data and processing burden of parallel operation. The results confirm that verifiable planning coupled with perception-aligned execution yields a near-optimal balance, delivering robust reliability with exceptional efficiency. \rev{The remaining performance gap stems primarily from inherent semantic ambiguities, where vague user descriptions preclude deterministic grounding to the static topology. Furthermore, perception limitations contribute to sporadic failures, as the underlying VLM faces challenges in identifying targets under complex real-world lighting or occlusion. }

\begin{table}[t]
\centering
\setlength{\abovecaptionskip}{0.03cm}
\caption{Real-World Scheduling Paradigm Comparison.}
\label{tab:paradigm_comparison_real}
\small

\begin{tabular}{@{}lcccc@{}}
    \toprule
    \textbf{Paradigm} & 
    \makecell[c]{\textbf{TCR}\\\textbf{(\%)}} & 
    \makecell[c]{\textbf{Latency}\\\textbf{(s)}} & 
    \makecell[c]{\textbf{Bandwidth}\\\textbf{(GB)}} & 
    \makecell[c]{\textbf{TFP}\\\textbf{(frames)}} \\
    \midrule
    Static Scheduling & 3.61 & 403.42 & 0.138 & 179 \\
    Naive Parallel & 65.64 & 927.99 & 0.540 & 704 \\
    \textbf{IoT-Brain (Ours)} & \textbf{49.84} & \textbf{413.69} & \textbf{0.131} & \textbf{166} \\ 
    \bottomrule
\end{tabular}
\vspace{-2mm}
\end{table}

\vspace{-0.03in}
\subsection{System Analysis and Sensitivity}
\label{subsec:sensitivity_real}

\begin{figure}[t]
    \centering
    \setlength{\abovecaptionskip}{0.03cm}
    \includegraphics[width=\columnwidth]{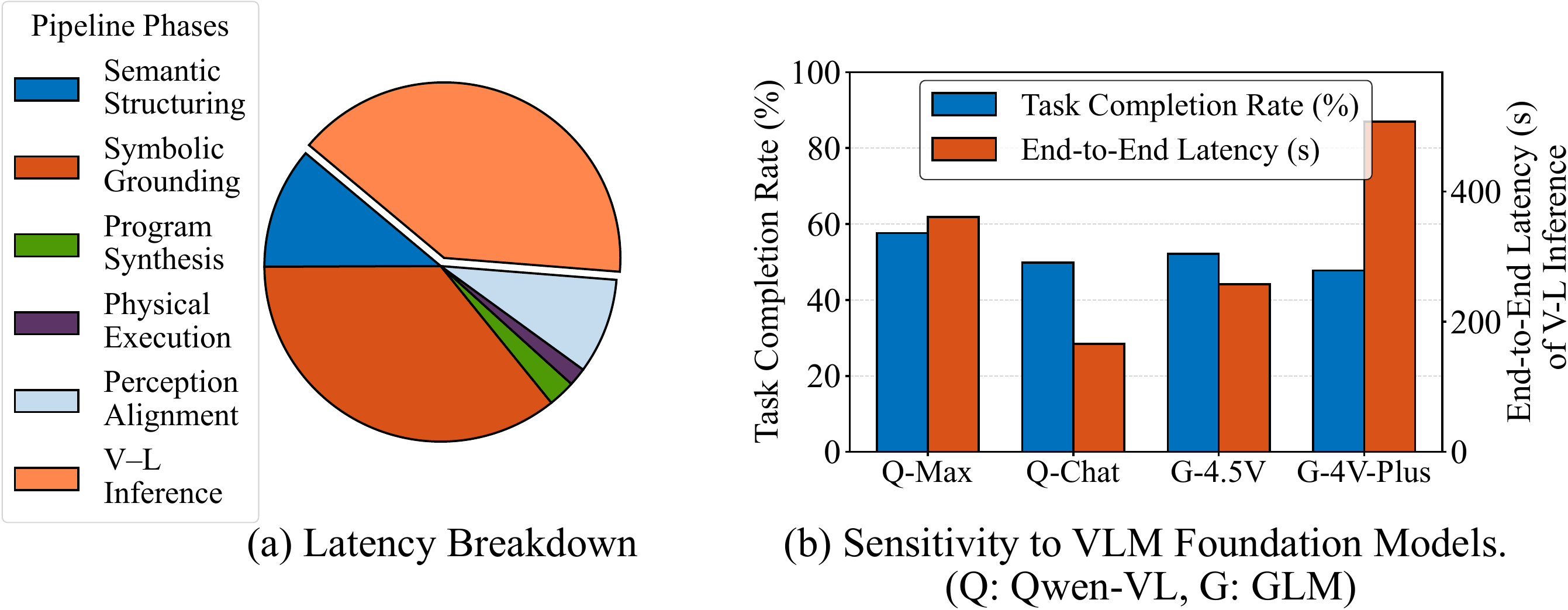} 
    \caption{System performance analysis.}
    \label{fig:sensitivity_vlm}
    \vspace{-2mm}
\end{figure}

\noindent\textbf{Latency Breakdown.} We profiled the end-to-end latency for the \textit{IoT-Brain} pipeline to characterize its temporal behavior. The breakdown in Fig.~\ref{fig:sensitivity_vlm}(a) clarifies the cost structure of complex semantic scheduling. The two dominant phases are \textit{Symbolic Grounding} and \textit{V–L Inference}, which account for most of the execution time on challenging Tier 2 \& 3 tasks. The high cardinality of scenarios necessitates extensive verification cycles during grounding, while a larger scheduled sensor set increases the volume of frames for VLM inference.
By contrast, the initial \textit{Semantic Structuring} stage is remarkably efficient.
Overall, this analysis reveals that the primary latency drivers are not inefficiencies in our planning engine, but rather the intrinsic complexity of the tasks, which demands substantial verification and perception effort.

\noindent\textbf{Sensitivity to VLM Foundation Models.} The framework’s end-to-end performance is naturally influenced by the specific capabilities of the VLM selected within the \textit{Perception Aligner}.
To demonstrate architectural generality, we evaluate our framework with a diverse spectrum of both local open-source and proprietary API-based VLMs.
As shown in Fig.~\ref{fig:sensitivity_vlm}(b), the results reveal a clear and quantifiable performance–latency trade-off.
Cloud-hosted API-based models such as Qwen-VL-Max achieve higher TCR but introduce significant network communication latency, whereas the lightweight open-source Qwen-VL-Chat offers lower latency with a corresponding reduction in TCR.
This empirical outcome underscores our framework's inherent robustness.
The \textit{STG} paradigm, by decoupling planning from perception, is fundamentally model-agnostic, providing a stable planning scaffold that functions effectively regardless of the underlying VLM.
The choice of foundation model thus becomes a configurable trade-off for deployers, allowing them to flexibly balance performance, cost, and data privacy.

\vspace{-0.02in}
\section{DISCUSSION}
\label{sec:discussion}

\noindent\textbf{Principles of Generalizability.}
\rev{
Although our current instantiation centers on a camera network, the \textit{STG} paradigm operates on a robust sensor-agnostic abstraction layer where nodes represent generic spatial coverage requirements rather than specific hardware interfaces.
The LLM reasoning engine deals exclusively with logical predicates (e.g., \texttt{is\_covered(location)}), while the adaptation to specific sensing modalities occurs entirely within the deterministic \textit{Verifying Toolkit}, which encapsulates the physical constraints.
For instance, replacing a camera with a microphone array or a thermal sensor only requires updating the toolkit's geometric calculation function to validate attributes like an acoustic detection radius or thermal sensing range instead of a visual field-of-view, while the high-level semantic planning logic remains structurally identical.
This modularity empowers the framework to extend to diverse IoT scenarios (e.g., audio sensing~\cite{Wang2024SLAMBasedJC} or thermal sensing~\cite{Motlagh2021HowLC}) without expensive retraining or fine-tuning of the core reasoning engine, though deployment in highly dynamic settings with frequent sensor outages would necessitate continual state updates~\cite{Krajnk2016PersistentLA}.
}

\noindent\textbf{System Deployment Effort.}
\rev{
\textit{IoT-Brain} is explicitly designed to streamline practical deployment and minimize engineering overhead through three strategic design choices.
First, regarding topological knowledge construction, our pipeline parses standard OSM data to semi-automatically generate the sensor-integrated topological representation (encoding attributes like FOV and sensing radius), limiting manual intervention primarily to the binding of specific sensor IDs.
Second, regarding model tuning, the framework is fundamentally training-free by leveraging the one-shot or few-shot prompting strategies detailed in \S\ref{sec:system_design}, thereby eliminating the need for expensive data collection or fine-tuning.
Third, regarding algorithmic integration, the system adopts a "composition over creation" approach by wrapping existing off-the-shelf solutions. Specifically, it invokes standard optimization algorithms (e.g., ILP solvers) via the API pool and integrates established models (e.g., YOLO, ReID) into the perception module, significantly reducing the overall development and maintenance complexity.
}

\noindent\textbf{Architectural Privacy-by-Design.} \rev{ Beyond simple policy compliance, privacy in \textit{IoT-Brain} is an inherent result of its decoupled architecture. First, we enforce a principle of symbolic isolation where the planning LLM, even if cloud-based, operates solely on abstract text symbols such as \texttt{Sensor\_01}. Crucially, the model never accesses raw privacy-sensitive sensor streams like video, audio, or thermal data, effectively creating a structural privacy air-gap. Second, the system ensures rigorous data minimization through its optimization objective. This mechanism mathematically enforces that sensors are activated only for the strictly necessary spatiotemporal windows verified by the graph, rather than performing invasive persistent monitoring. This structural guarantee remains valid regardless of the sensing modality, providing a robust blueprint for privacy-preserving intelligent sensing aligned with strict organizational requirements~\cite{Magara2024InternetOT,Meneghello2019IoTIO}. }

\vspace{-0.12in}
\section{RELATED WORK}
\label{sec:related_work}

\noindent\textbf{LLMs for Sensor System Control.} Recent pioneering works have begun to explore using LLMs to translate high-level human intent into executable sensor actions, typically operating within strictly constrained smart-home settings~\cite{king2024sasha, Liu2025TaskSenseAT, AlSafi2025VegaLI}. To mitigate the model's unpredictability, these approaches often rely on formal grammars or predefined API templates to enhance plan robustness. However, while such methods ensure syntactic validity, they fail to address the complexity of large-scale spatial reasoning. Our work differs fundamentally by tackling the twin challenges of campus-scale scalability and physical-world grounding. We shift the research focus from ensuring small-scale execution robustness to achieving verifiable, resource-optimal scheduling in massive deployments, where resource contention and topological constraints are paramount.

\noindent\textbf{LLM-based Agentic Planning.}
The advent of LLMs has catalyzed the development of sophisticated agentic planning frameworks, most notably including the Hierarchical~\cite{Shen2023HuggingGPTSA,Ge2023OpenAGIWL}, Reactive~\cite{Yao2022ReActSR,Shinn2023ReflexionLA}, and Backtracking~\cite{Yao2023TreeOT,Qin2023ToolLLMFL} paradigms.
These approaches excel at reasoning within reliable digital substrates, which are typically exemplified by software APIs or coding environments where execution feedback is immediate and deterministic.
However, such methods lack the mechanisms to handle the ambiguity inherent in the physical world.
Our work centers on grounding language-based reasoning in this noisy reality.
To achieve this, \textit{STG} functions as a rigorous neurosymbolic scaffold that constrains the LLM’s reasoning process through verifiable graph construction.
This architecture supplies the essential guarantees needed for reliability in high-stakes embodied settings.
 
\noindent\textbf{Language-Guided Perception.}
Historically, research in person re-identification has centered primarily on metric learning for image-to-image retrieval~\cite{Zhou2019OmniScaleFL, Zhou2019LearningGO, Chen2017PersonRB}.
The recent introduction of large language models has enabled powerful cross-modal alignment, facilitating text-to-image re-identification and language-guided tracking in continuous video streams~\cite{Li2024LaMOTLM,Wu2023ReferringMT}.
Yet, most such studies presume a predefined, passive sensor stream or an exhaustively broad search space.
Our \textit{Perception Aligner} fundamentally challenges this passive paradigm by operating within a proactively scheduled, on-demand feed.
It provides just-in-time verification to drive the next sensor activation, ensuring the system performs efficient, closed-loop active perception rather than relying on persistent, resource-intensive, and often unscalable open-domain tracking mechanisms.

\vspace{-0.12in}
\section{CONCLUSION}
\label{sec:conclusion}

We formalize \textit{Semantic-Spatial Sensor Scheduling (S³)} and reveal critical gaps in LLM planning. To address these, we introduce \textit{STG}, a "verify-before-commit" neurosymbolic paradigm that decouples semantic inference from deterministic scheduling.
Our implementation, \textit{IoT-Brain}, evaluated on our \textit{TopoSense-Bench} benchmark and real-world deployment, demonstrates significant gains in both reliability and efficiency over representative planners. These contributions provide a practical blueprint for LLMs to translate high-level intent into correct physical action robustly.

\vspace{-0.1in}
\begin{acks}
We thank the anonymous MobiCom reviewers and shepherd for their constructive comments. This research was supported by the China National Natural Science Foundation with No. 623B2093, No. 62441228 and Science and Technology Tackling Program of Anhui Province No.202423k09020016.
\end{acks}

\input{main.bbl}


\end{document}